 \documentclass[sigconf]{acmart}
\usepackage{multirow}

\AtBeginDocument{%
  \providecommand\BibTeX{{%
    \normalfont B\kern-0.5em{\scshape i\kern-0.25em b}\kern-0.8em\TeX}}}

\setcopyright{none}

\acmConference[1st HEAL Workshop at CHI Conference on Human Factors in Computing Systems]{}{May 12}{Honolulu, HI, USA}

\acmDOI{}

\acmISBN{}

\begin{document}

\title[Identifying Misleading Headlines with LLM]{Exploring the Potential of the Large Language Models (LLMs) in Identifying Misleading News Headlines}

\author{Md Main Uddin Rony}
\affiliation{%
  \institution{University of Maryland}
  \city{College Park}
  \state{Maryland}
  \country{USA}
}
\email{mrony@umd.edu}

\author{Md Mahfuzul Haque}
\affiliation{%
  \institution{University of Maryland}
  \city{College Park}
  \state{Maryland}
  \country{USA}
}
\email{mhaque16@umd.edu}

\author{Mohammad Ali}
\affiliation{%
  \institution{University of Maryland}
  \city{College Park}
  \state{Maryland}
  \country{USA}
}
\email{mali24@umd.edu }

\author{Ahmed Shatil Alam}
\affiliation{%
  \institution{University of Oklahoma}
  \city{Norman}
  \state{Oklahoma}
  \country{USA}
}
\email{asalam@ou.edu}

\author{Naeemul Hassan}
\affiliation{%
  \institution{University of Maryland}
  \city{College Park}
  \state{Maryland}
  \country{USA}
}
\email{nhassan@umd.edu}

\renewcommand{\shortauthors}{Rony et al.}

\begin{abstract}
  In the digital age, the prevalence of misleading news headlines poses a significant challenge to information integrity, necessitating robust detection mechanisms. This study explores the efficacy of Large Language Models (LLMs) in identifying misleading versus non-misleading news headlines. Utilizing a dataset of 60 articles, sourced from both reputable and questionable outlets across health, science \& tech, and business domains, we employ three LLMs—ChatGPT-3.5, ChatGPT-4, and Gemini—for classification. Our analysis reveals significant variance in model performance, with ChatGPT-4 demonstrating superior accuracy, especially in cases with unanimous annotator agreement on misleading headlines. The study emphasizes the importance of human-centered evaluation in developing LLMs that can navigate the complexities of misinformation detection, aligning technical proficiency with nuanced human judgment. Our findings contribute to the discourse on AI ethics, emphasizing the need for models that are not only technically advanced but also ethically aligned and sensitive to the subtleties of human interpretation.
\end{abstract}

\settopmatter{printfolios=true}
\maketitle

\vspace{-1em}

\section{Introduction}
News headlines are precursors to comprehensive stories and serve as persuasive messages, making their accuracy and authenticity crucial. Gabielkov et al. note that many readers may not proceed beyond the headlines to read the full content~\cite{gabielkov2016social}; however, they can still receive misleading information if these headlines do not accurately represent the content. We use the term Misleading News Headlines to describe this particular phenomenon. Misleading News Headlines arise when the headline of a news article fails to represent its content accurately. Consider the following example for illustration.

\begin{quote}
    \textbf{Headline: } Hot tea linked to increased risk of esophageal cancer~\footnote{https://tinyurl.com/misleading-headline-example1} \\
    \textbf{Content:} People who like hot tea may want to wait until it gets cooler before taking that first sip. Drinking more than 700 milliliters of tea at higher than 60 degrees Celsius, or 140 degrees Fahrenheit, was linked to a 90 percent increased risk of esophageal cancer, according to a study ...
    
    ``Many people enjoy drinking tea, coffee, or other hot beverages. However, according to our report, drinking very hot tea can increase the risk of esophageal cancer," said lead author Farhad Islami, a researcher at the American Cancer Society and study lead author, in a news release. ... In 2016, the International Agency for Research on Cancer said that drinking any drink over 65 degrees Celsius makes it a carcinogen or something likely to cause cancer.
    Other studies have linked drinking hot tea and drinking excessive amounts of alcohol daily to esophageal cancer, as well.
\end{quote}

The headline \textit{Hot tea linked to increased risk of esophageal cancer} is misleading because it specifically singles out hot tea, despite the article indicating that the risk is associated with consuming any very hot beverage. This narrow focus on hot tea could lead readers to incorrectly believe that only hot tea poses this cancer risk, potentially causing them to overlook the similar risks associated with other hot beverages. Consequently, readers might make uninformed decisions about their beverage choices, erroneously assuming that switching from hot tea to another hot drink, like coffee, would mitigate their risk of esophageal cancer when the temperature, not the type of beverage, is crucial.

 If the headline is misleading, it may cause a wrong impression, leading to uninformed decision-making~\cite{ecker2014effects}. Addressing the issue of Misleading News Headlines is critical to rebuilding trust in journalism and combating misinformation. Manual evaluations, while effective, are impractical due to the sheer volume and speed of news dissemination, necessitating automated solutions. However, constructing such systems presents challenges, particularly the need for extensive, high-quality data. Large-scale, representative datasets are essential for training robust machine learning models. This complexity underscores the significance of leveraging advanced techniques like Large Language Models (LLMs) to detect and classify misleading headlines accurately, ultimately enhancing journalism's credibility and its ability to counteract misinformation.

Recent advances in natural language processing have led to powerful Large Language Models (LLMs) capable of understanding complex languages intricately ~\cite{min2023recent}. These LLMs have been successfully applied to identify and rectify vaccine misinformation, showcasing their potential in public health communication and information validation ~\cite{deiana2023artificial}. However, it's essential to acknowledge that misleading headlines differ from other forms of misinformation. Misleading headlines often straddle a fine line, potentially presenting skewed or exaggerated information without being entirely false. This distinct nature adds complexity to the task of utilizing LLMs to detect and address misleading headlines accurately. In light of this gray area, designing detection mechanisms can be more complex, resulting in the following research question: \\RQ: To what extent can Large Language Models accurately identify headlines as misleading?

\section{Related Work}
\subsection{Overview of Misleading Headline Detection}
Misleading headlines create a disconnect between the title and the article's content, leading to potential misinterpretation by readers. These headlines may present overrated, false, or unsupported information, aiming to attract attention or drive web traffic through exaggerated or sensational content \cite{chesney2017incongruent, yoon2019detecting}. They often leverage emotional language, biasing readers even before they engage with the article, and may omit key information or emphasize less relevant details, leading to confusion and misinformation \cite{wei2017learning, ecker2014effects}. The challenge lies in the headlines' ability to reinforce existing beliefs, making misinformation appear more credible and difficult to correct, thus significantly impacting reader understanding and opinion formation \cite{ecker2015misinfopsycho, piksa2022cognitive}.
A key challenge highlighted in the literature for automated misleading headline detection is the inadequacy of existing datasets and NLP methods in capturing incongruence between headlines and article content. This gap necessitates the development of more nuanced datasets and methodologies that go beyond simple agreement or disagreement models~\cite{chesney2017incongruent}. Additionally, the variability in dataset creation methods and the limitations of existing datasets in representing the full scope of misleading headlines pose significant challenges to developing effective automated detection systems~\cite{mishra2020musem,park2020baitwatcher}.

\subsection{Overview of LLMs in NLP and Misinformation}
The proliferation of Large Language Models (LLMs) in natural language processing represents a significant leap forward, enabling these models to grasp complex language structures with remarkable depth ~\cite{min2023recent}. Demonstrated by their effectiveness in combating vaccine misinformation, LLMs hold promise for enhancing public health communication and ensuring the accuracy of information ~\cite{deiana2023artificial}. Nonetheless, the challenge of misleading headlines, which may convey skewed or exaggerated information without being outright false, underscores a unique dilemma. This subtlety complicates the use of LLMs for detecting and addressing misleading content, revealing a gap in their application. This nuanced challenge underlines the need for research into the capabilities of LLMs to discern and classify misleading headlines, highlighting a critical area for exploration.

\section{Method}
\subsection{Data Collection}
In our study, we collected news articles from 12 sources, categorized into reliable (e.g., ABC News, NY Times, Washington Post) and unreliable (e.g., Infowars, Lifezette) groups based on assessments from Media Bias/Fact Check (MBFC)~\footnote{https://mediabiasfactcheck.com/}, a third-party website that evaluates media source credibility. Our focus was on articles within the Health, Science \& Tech, and Business domains. Three domain-knowledgeable annotators selected five articles from each domain from four sources, starting from March 31st, 2022, assessing if headlines were misleading by reviewing both the headline and the content. This process yielded a balanced preliminary dataset of 60 articles, comprising 30 misleading and 30 non-misleading headlines.

During annotation, each annotator independently reviewed 40 articles (20 misleading and 20 non-misleading) compiled by the others, without source identifiers to avoid bias. The review process involved three rounds of detailed examination to label articles as misleading or non-misleading, with annotations reflecting varying confidence levels. Consensus was reached on 18 articles being unanimously misleading, while at least two annotators agreed on 27 articles. Given our rigorous criterion that a headline is considered misleading if it could potentially mislead at least one reader, the final dataset consisted of 37 misleading and 23 non-misleading headlines.
\vspace{-1em}
\subsection{LLM Evaluation}
ChatGPT(version 3.5 and 4) and Gemini evaluated the collected headlines for labels and explanations, aiming to understand their capability to identify misleading headlines.

The headlines and relevant news content were submitted to LLMs for assessment. The LLMs determined if the headline is misleading and explain their decisions. The API requests were sent to the LLMs, which evaluated the news content's representation and provided a decision and an explanation.A sample request prompt would be as follows:
\begin{quote}
\textit{prompt=``Evaluate if the following headline is misleading based on the news content provided. 
Headline: [Your Headline Here]
News Content: [Your News Content Here]
Is this headline misleading? Please explain your decision."}
\end{quote}

\section{Results}
This study aimed to assess the capability of large language models (LLMs) — specifically ChatGPT-3.5, ChatGPT-4, and Gemini — to detect and explain misleading news headlines accurately. Employing a dataset of 60 news articles, where human annotators identified 37 as having misleading headlines, we explored how these LLMs could align with human judgment in identifying misinformation.

\subsection{LLM's Classification Performance Analysis}
This section presents the findings from evaluating three Large Language Models (LLMs) — ChatGPT-3.5, ChatGPT-4.0, and Gemini based on a binary classification task. 

\subsubsection{LLMs' Overall Performance Analysis}
Each LLM was assessed through precision, recall, f1-score, and overall accuracy metrics, providing insight into their effectiveness in addressing the RQ1.

\begin{table}[]
\centering
\caption{Performance of LLMs in Detecting Misleading News Headlines}
\label{table:LLMPerformanceReorganized}
\resizebox{\columnwidth}{!}{
\begin{tabular}{|l|ccc|ccc|c|}
\hline
\textbf{Model} & \multicolumn{3}{c|}{\textbf{Non-misleading}} & \multicolumn{3}{c|}{\textbf{Misleading}} & \textbf{Accuracy} \\
 & \textbf{Precision} & \textbf{Recall} & \textbf{F1} & \textbf{Precision} & \textbf{Recall} & \textbf{F1} &  \\
\hline
ChatGPT-3.5 & 1.00 & 0.09 & 0.16 & 0.46 & 1.00 & 0.63 & 0.48 \\
ChatGPT-4   & 0.85 & 0.97 & 0.90 & 0.95 & 0.77 & 0.85 & 0.88 \\
Gemini      & 0.68 & 0.79 & 0.73 & 0.65 & 0.50 & 0.57 & 0.67 \\
\hline
\end{tabular}
}
\end{table}

\textbf{ChatGPT-3.5 Performance} The performance of ChatGPT-3.5 showed a high level of precision in identifying non-misleading headlines (precision: 1.00) but with a notably low recall rate (recall: 0.09), indicating a tendency to misclassify non-misleading headlines as misleading. Conversely, for misleading headlines, the model demonstrated a lower precision (0.46) but a perfect recall score (1.00), suggesting it was effective in identifying misleading headlines but with a considerable rate of false positives. The accuracy of ChatGPT-3.5 stood at 48\%, with a macro-average f1-score of 0.39, indicating a moderate level of imbalance in its classification capability, skewed towards identifying misleading headlines.

\textbf{ChatGPT-4.0 Performance} ChatGPT-4.0 significantly improved over its predecessor, achieving an accuracy of 88\%. It showed high precision and recall in identifying both misleading (precision: 0.95, recall: 0.77) and non-misleading headlines (precision: 0.85, recall: 0.97), reflected in a balanced f1-score for non-misleading (0.90) and misleading (0.85) headlines. The macro and weighted average f1-scores were both close to 0.88, illustrating a robust capability in accurately classifying headlines while maintaining a balanced performance across both classes.

\textbf{Gemini Performance} Gemini's performance presented a balanced approach between the two extremes of ChatGPT-3.5 and ChatGPT-4.0, with an overall accuracy of 67\%. It demonstrated moderate precision and recall for non-misleading (precision: 0.68, recall: 0.79) and misleading headlines (precision: 0.65, recall: 0.50), leading to an f1-score of 0.73 and 0.57, respectively. The macro and weighted average f1-scores were 0.65 and 0.66, indicating a reasonable but not optimal balance in classification capability across the two categories.

\subsubsection{LLM's Performance by Consensus Level}
The efficacy of LLMs in identifying misleading content was examined in contexts of unanimous consensus by annotators versus mixed consensus (Majority and Minority Misleading) ( See in Table ~\ref{tab:performance-consensus}).

\textbf{Unanimous Consensus} In scenarios where human annotators unanimously agreed on the nature of the headlines (either misleading or not misleading), ChatGPT-4 exhibited the highest performance, accurately classifying misleading headlines with an accuracy of 83.3\% and non-misleading headlines with 95.7\%. Gemini followed with 61.1\% accuracy for misleading and 73.9\% for non-misleading headlines. ChatGPT-3.5 showed a topmost accuracy, with 94.4\% for misleading but performed poorly for non-misleading headlines with 8.7\% accuracy. These results indicate a potential alignment between advanced LLM judgments and unanimous human consensus.

% \usepackage{multirow}
% \usepackage{graphicx}
\begin{table}[]
\caption{LLMs' Performance by Human Consensus Level}
\label{tab:performance-consensus}
\resizebox{\columnwidth}{!}{%
\begin{tabular}{|l|l|c|r|}
\hline
\multicolumn{1}{|c|}{\textbf{Consensus level}} & \textbf{Model}               & \textbf{Is Misleading?} & \textbf{\# of Headlines} \\ \hline
\multirow{6}{*}{Unanimous Not Misleading}      & \multirow{2}{*}{Gemini}      & Yes                     & 6                        \\ \cline{3-4} 
                                               &                              & No                      & 17                       \\ \cline{2-4} 
                                               & \multirow{2}{*}{ChatGPT-4}   & Yes                     & 1                        \\ \cline{3-4} 
                                               &                              & No                      & 22                       \\ \cline{2-4} 
                                               & \multirow{2}{*}{ChatGPT-3.5} & Yes                     & 21                       \\ \cline{3-4} 
                                               &                              & No                      & 2                        \\ \hline
\multirow{6}{*}{Minority Misleading}           & \multirow{2}{*}{Gemini}      & Yes                     & 2                        \\ \cline{3-4} 
                                               &                              & No                      & 8                        \\ \cline{2-4} 
                                               & \multirow{2}{*}{ChatGPT-4}   & Yes                     & 2                        \\ \cline{3-4} 
                                               &                              & No                      & 8                        \\ \cline{2-4} 
                                               & \multirow{2}{*}{ChatGPT-3.5} & Yes                     & 9                        \\ \cline{3-4} 
                                               &                              & No                      & 1                        \\ \hline
\multirow{6}{*}{Majority Misleading}           & \multirow{2}{*}{Gemini}      & Yes                     & 2                        \\ \cline{3-4} 
                                               &                              & No                      & 7                        \\ \cline{2-4} 
                                               & \multirow{2}{*}{ChatGPT-4}   & Yes                     & 3                        \\ \cline{3-4} 
                                               &                              & No                      & 6                        \\ \cline{2-4} 
                                               & \multirow{2}{*}{ChatGPT-3.5} & Yes                     & 8                        \\ \cline{3-4} 
                                               &                              & No                      & 1                        \\ \hline
\multirow{6}{*}{Unanimous Misleading}          & \multirow{2}{*}{Gemini}      & Yes                     & 11                       \\ \cline{3-4} 
                                               &                              & No                      & 7                        \\ \cline{2-4} 
                                               & \multirow{2}{*}{ChatGPT-4}   & Yes                     & 15                       \\ \cline{3-4} 
                                               &                              & No                      & 3                        \\ \cline{2-4} 
                                               & \multirow{2}{*}{ChatGPT-3.5} & Yes                     & 17                       \\ \cline{3-4} 
                                               &                              & No                      & 1                        \\ \hline
\end{tabular}%
}
\end{table}

\textbf{Mixed Consensus (Majority \& Minority Misleading)}
\begin{itemize}
    \item Majority Misleading: When a majority (but not all) of the annotators identified headlines as misleading, ChatGPT-4's performance significantly decreased to 33.33\% accuracy for misleading headlines. While Gemini experienced a more pronounced drop to 22.2\%, ChatGPT-3.5 demonstrated a better performance with an accuracy rating of 88.9\%, which is generally due to the tendency to misclassify non-misleading headlines as misleading. The results of this study suggest that there may be challenges in cases where there is less clear-cut human agreement.
    \item Minority Misleading: For headlines deemed misleading by a minority of annotators, ChatGPT-4's accuracy was 20\%. Although Gemini exhibited the same accuracy as ChatGPT-4, ChatGPT-3.5 performed significantly better (90\%) than its counterpart models, which underscores the difficulty LLMs have when there is a lack of strong human consensus.
\end{itemize}

% \subsection{LLM's Explanation Analysis}

\section{Discussion}

The evaluation of Large Language Models (LLMs) in distinguishing misleading news headlines reveals essential insights into the intersection of artificial intelligence and media integrity. This discussion delves into the implications of our findings within the broader context of human-centered evaluation and auditing methods for LLMs, highlighting the nuanced role these models play in supporting stakeholders across the digital information landscape.

\subsection{Integrating Human-Centered Evaluation in LLM Auditing}
As presented in our findings, the exploration of the effectiveness of large language models (LLMs) in discerning misleading news headlines emphasizes the imperative for incorporating human-centered evaluation and auditing frameworks. This approach not only benchmarks the performance of LLMs against human judgment but also aligns with the broader discourse on enhancing AI interpretability and reliability in media contexts~\cite{diakopoulos2017algorithmic}.
\vspace{-1em}
\subsection{LLM Performance and Human Consensus}
\subsubsection{Alignment with Unanimous Consensus}
The better performance of ChatGPT-4 in instances of unanimous human consensus on misleading headlines highlights the advancements in AI's capability to parallel human reasoning in clear-cut scenarios. This observation resonates with the literature emphasizing the need for AI systems to understand and replicate human-like judgment in tasks requiring nuanced interpretation ~\cite{rudin2019stop}. Such alignment is crucial for stakeholders, including media professionals and content moderators, who rely on AI to filter through vast amounts of data for potential misinformation.
\vspace{-1em}
\subsection{Navigating Mixed Consensus}
The nuanced challenge presented by mixed human consensus highlights a frontier in AI development. The differential performance of LLMs, particularly in majority and minority misleading scenarios, reflects the complexity of human cognition and the subjective nature of misinformation. This observation aligns with the AI ethics community's push for models that are not just technically advanced but also attuned to the nuances of human thinking and ethical concerns~\cite{rahwan2019machine,whittlestone2019role}.
\vspace{-1em}
\subsection{Implications for Stakeholders}
The practical implications of these findings are manifold. For journalists and media outlets, the deployment of LLMs that accurately identify misleading headlines could represent a significant step forward in maintaining informational integrity. For developers and AI researchers, our study highlights the importance of embedding human-centered design principles in the development of LLMs, ensuring these tools are both effective and ethically aligned with societal norms~\cite{shneiderman2020human}.

Moreover, for policymakers and regulators, understanding the capabilities and limitations of LLMs in identifying misinformation is crucial for crafting guidelines that promote responsible AI use in journalism and beyond. This aligns with ongoing discussions about the regulatory frameworks necessary to govern AI's application in sensitive societal domains~\cite{dignum2019responsible}.
\vspace{-1em}
\subsection{Future Research Direction}
Future research should aim to bridge the gap between LLM performance and the diverse ranges of human judgment, particularly in ambiguous or controversial scenarios. This includes investigating methodologies for incorporating ethical reasoning and bias recognition into LLM training processes. Additionally, expanding the scope of LLM training to encompass multimodal content could enhance their applicability across various media formats, offering a more holistic approach to misinformation detection.

A critical area for future exploration is the examination of explanations generated by LLMs in identifying misleading headlines and how these explanations align with human rationale. Understanding the logic and reasoning behind LLM decisions is essential for improving their reliability and trustworthiness. Analyzing LLM-generated explanations can provide insights into the models' interpretive processes, identifying areas where they may diverge from human thought patterns. This line of inquiry not only contributes to the development of more sophisticated and human-like LLMs but also supports the creation of AI systems whose decision-making processes are transparent, explainable, and, most importantly, aligned with ethical standards and societal expectations.
\vspace{-1em}
\section{Conclusion}
Our investigation into the capabilities of Large Language Models (LLMs) to identify misleading news headlines highlights the potential and challenges inherent in aligning AI with human judgment and ethical considerations. The study reveals that while models like ChatGPT-4 show promise in closely mirroring human decisions, particularly in clear-cut cases, discrepancies in performance across varying levels of human consensus highlight the complexity of misleading headline detection. The findings advocate for a human-centered approach in the development and evaluation of LLMs, emphasizing the need for models that are not only technically adept but also sensitive to the nuances of human ethics and reasoning. Future research directions, including examining LLM-generated explanations and expanding training to multimodal content, promise to further bridge the gap between AI and human judgment, paving the way for more reliable, ethical, and effective tools in combating misinformation.

\bibliographystyle{ACM-Reference-Format}
\bibliography{main}

\end{document}